\documentclass[letterpaper, 10 pt, conference]{ieeeconf}

\IEEEoverridecommandlockouts
\usepackage[utf8]{inputenc}
\usepackage{amsmath}

\usepackage{amsfonts}
\usepackage{amsthm}
\usepackage{amssymb}
\usepackage{graphicx}
\usepackage{booktabs}
\usepackage{url}

\usepackage{xcolor}

\usepackage{algorithm}
\usepackage{algpseudocode}

\title{\textbf{Bidirectional Communication Control for Human-Robot Collaboration}} 
\author{Davide Ferrari, Federico Benzi and Cristian Secchi

\thanks{D. Ferrari, F. Benzi and C. Secchi are
    with the Department of Sciences and Methods of Engineering,
    University of Modena and Reggio Emilia, Italy  {\tt\small
    \{davide.ferrari95,federico.benzi,
    cristian.secchi\}@unimore.it}}} 
      
\begin{document}

\maketitle

\begin{abstract}
    
    
    A fruitful collaboration is based on the mutual knowledge of each other skills and on the possibility of communicating their own limits and proposing alternatives to adapt the execution of a task to the capabilities of the collaborators. This paper aims at reproducing such a scenario in a human-robot collaboration setting by proposing a novel communication control architecture. Exploiting control barrier functions, the robot is made aware of its (dynamic) skills and limits and, thanks to a local predictor, it is able to assess if it is possible to execute a requested task and, if not, to propose alternative by relaxing some constraints. The controller is interfaced with a communication infrastructure that enables human and robot to set up a bidirectional communication about the task to execute and the human to take an informed decision on the behavior of the robot. A comparative experimental validation is proposed.
    
    
    
\end{abstract}

\section{Introduction}\label{sec:intro}

    
    Human-Robot Collaboration (HRC) is rapidly gaining importance in modern society. More and more applications in the industrial \cite{Vil18Mechatronics} \cite{Vys16mmscience}, medical \cite{Kap16ICRA} \cite{Bau16IEEE} \cite{Mis20IEEE}, elder assistance \cite{Bro11AAAI} \cite{Mas15AAL} and domestic \cite{Kid08ICIRS} \cite{Lee05IEEE} sectors require robots to work alongside people as capable members of human-robot teams. Within these teams, just as it happens in human-human collaboration (HHC), the awareness of one's abilities and limits, the analysis of the environment and the knowledge of the intentions of the user assume fundamental importance. Furthermore, the communication between team members plays a crucial role in sharing such information, necessary for the correct performance of the task into the shared workspace and reporting the intentions or requests to the other party of the team \cite{Bou12Neurorobotics}.
    In the literature multiple communication and control strategies have tried to manage teamwork operations: in \cite{LIU2020CIRP}, the authors presented a multimodal control architecture for symbiotic HRC assembly driven by voice instructions, hand motion recognition and body motion commands. In \cite{Qui2015ROMAN} it developed a two-way visual pointing gesture communication architecture that leverages a leader-follower relationship to guide the robot to the correct positions during a pick and place task. Similarly in \cite{Ros2020IROS} it showed a decision-theoretic model in which a robot interprets multimodal human communication to disambiguate item references by asking questions.
    Other works have preferred to overshadow communication and focus on analyzing the work area and prediction human intentions, e.g. by using proactive systems of recognition to infer hidden user intentions \cite{Sch2005ROMAN} or predictive and anticipatory models such as Recurrent Neural Network Prediction \cite{Sch2018ICRA} or Partially Observable Markov Decision Process (POMDP) \cite{Kar2010HRI}.
    As we can see in the examples above, teamwork communication in HRC rarely plays a fundamental role, but is often replaced by predictive algorithms to make robotic systems as autonomous as possible. 
    Even when communication is in the foreground, it is almost always guided by the user, providing the robot with predefined commands and prompts which simply executes them or gives a negative response in case of failure.
    In yet other cases, communication is just  used as feedback to inform the user about the intentions of the robot \cite{Gru2021Sensors}, excluding the possibility of a two-way exchange of intentions. 
    
    These results are very different from what happens during a human-to-human interaction, in which both team members exchange information and make requests, evaluating their feasibility, the occurrence of possible errors or risks and propose possible alternative solutions if required. In order to mimic/reproduce this, the robot must be aware of its own skills, capabilities and limits and able to dynamically adapt to the surrounding environment, which is constantly changing during the operations. Consequently, the control strategy assumes fundamental importance: it must be capable of dynamically adapting its behavior, changing it online according to both the current conditions of the workspace, both the limitations proper of the robotic application. Additionally, it should foresee the evolution of each task over time and coordinate with the operator based on the estimated outcomes. Control Barrier Functions (CBFs) \cite{ames2019control} lend themselves particularly well from this perspective, as they have been successfully exploited for dynamically constraining the behavior of the robot \cite{Notomista2018CoRR, Zha20toh}. In particular, Time Varying CBFs \cite{Not18ral} have been deployed for ensuring safety in HRC scenarios \cite{Fer20ras, Fer20ral} as well as multiple task execution \cite{notomista2020settheoretic, benzi2021optimal}, thus providing a formulation of the controller that fits our previously stated needs.

    The architecture presented in this paper approaches HRC in a new way: robot and human are considered as equally capable members, as is the case in human-human collaboration, allowing for a bidirectional communication channel that ensures effective, fast and intuitive communication among them. Both team members can exchange information useful for coordinating the job, such as updating on the current status of the task or communicating their intentions, and they can make requests for help and cooperation (e.g. ask for a tool, screw a bolt) or explanations of additional information necessary to complete the task (e.g. speed of execution, position of a desired object). 
    
    Thus, the contributions of this paper are:
    \begin{itemize}
        \item A novel Bidirectional Communication Control Architecture between human and robot in HRC.
        \item A control-prediction structure that enables the robot to evaluate commands, generate alternatives and propose them to the operator in order to reach a mutual agreement in terms of task execution.
        \item An experimental validation of the architecture by comparing bidirectional communication control and one-way common interaction during an HRC assembly task.
    \end{itemize}
    
     
    This paper is organized in the following way: in Section \ref{sec:proposed architecture}, we introduce the proposed architecture. In Sections \ref{sec:communication}, \ref{sec:control} and \ref{sec:decision}, we describe the communication, control-prediction and decision layers. In Section \ref{sec:exp}, we present the architecture implementation and analyze the experimental results. We conclude and outline some ideas for future works in Section \ref{sec:conclusions}.

\vspace{-2mm}
\section{Proposed Architecture}\label{sec:proposed architecture}

    
    The goal of this paper is to develop a Bidirectional Communication Control Architecture that makes it possible to create a two-way communication between human and robot, in order to ensure greater efficiency during the performance of HRC tasks.
    The proposed architecture must be able to collect the inputs of the user, analyze them to extrapolate any requests, communications or questions and process them with a decision making system that will have to provide a response to the user, depending on the nature of the request and the feedback received from the control algorithm on the status of the robot and the surrounding environment. Moreover, the system must be conscious of its capabilities and limits and aware of the external environment in order to guarantee safety during teamwork, communicate the feasibility of the desired operations and propose acceptable alternatives and compromises to carry out the most difficult requests.
    
    
    \begin{figure} [htbp]
        \centering
        \vspace{-3mm}
        \includegraphics[trim=3.2cm 2.1cm 2.2cm 2.5cm,clip,scale=0.25]{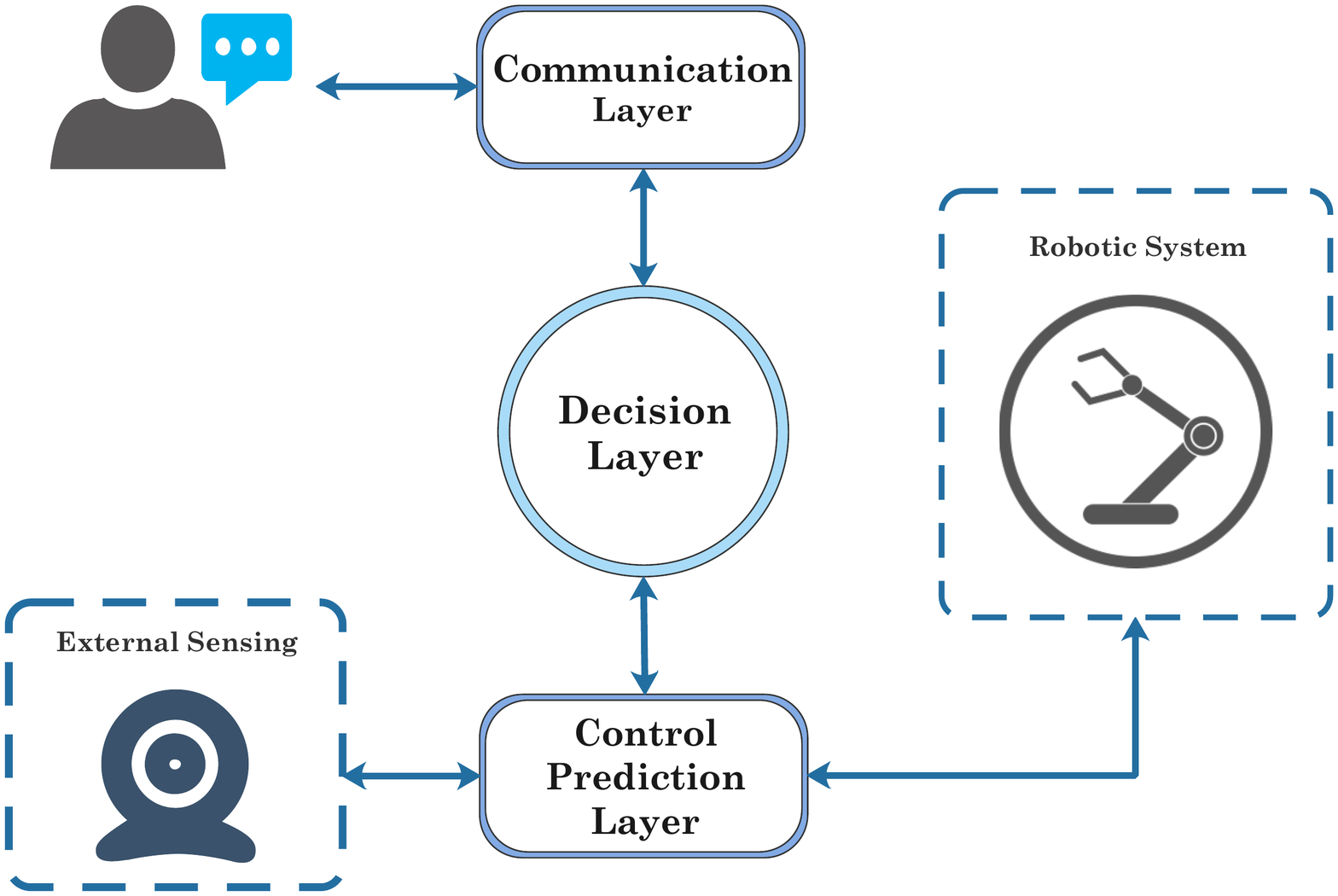}
        \caption{Overall Architecture}
        \label{fig:Overall Architecture}
        \vspace{-1mm}
    \end{figure}
    
    
    The overall architecture, in Fig. \ref{fig:Overall Architecture}, consists of three layers: a communication layer that has the task of managing the exchange of information with the operator, a control-prediction layer for enabling the robot to communicate contents based on its skills/limits and a decision layer to manage the flow of information and choose the action or response that the robot will return.
    The communication layer, through a generic communication channel (e.g. command shell, touchscreen, voice, gesture), is responsible for acquiring the information that the user wants to provide to the robotic system and returning the processed response. This information is then sent to the decision layer where the actions of the robot and the responses to be returned to the user will be processed based on the nature of the requests, the external environment and the robot feedback coming from the control layer. The latter is in charge of computing the low level control input for the robot by means of a CBF-based optimization framework, which takes into account all the information regarding the task, the limits and capabilities of the robot, as well as the current condition of the working scenario, for producing the best possible input.
    The exchange of information between these three levels makes it possible to process complex interactions between human and robot during HRC tasks, exploiting natural and effective communication.

\section{Communication Layer}\label{sec:communication}

    The communication layer has the task of allowing the exchange of information between the operator and the robotic system: it must be able to acquire the intention of the operator (in the form of sentences, gestures, interface inputs, etc.), process them to extrapolate their meaning and convert them into useful information for the decision layer. It will also need to be capable of processing an appropriate response based on the decision layer output and communicating it to the operator. There are multiple solutions to build a communication channel between humans and robots, ranging from the simplest such as command line interfaces, push button panels and sound and light signals to more complex architectures such as interactive monitors, mobile devices, voice interfaces and optical devices. In the architecture proposed in this article it has been chosen to use a commercial voice assistant to implement a vocal communication channel, bringing out several advantages:
    
    \begin{itemize}
        \item Voice is one of the most used communication channels within a human-human teamwork and manages to guarantee a natural and spontaneous interaction.
        \item Using a commercial voice assistant provides the possibility to exploit already existing linguistic models, achieving high reliability and precision.
        \item The creation of a voice module that is easy to implement in the robotic environment and with a very low purchase and use cost, allows to create a communication channel that can be integrated into a wide variety of systems.
    \end{itemize}
    

    
    The architecture of the communication layer
    consists of two main components: the Skill Server and the Text-To-Speech (TTS) blocks. The Skill Server block has the task of putting the robotic ecosystem in communication with the custom application of the voice assistant, called Skill or Action. The TTS block, on the other hand, has the task of reproducing, through the voice assistant, the communications or requests that come directly from the robotic system.
    
    
    
    The voice assistant records the communication of the operator and process the audio data stream on the servers in the cloud, extrapolating the intentions of the user and building a request to invoke the skill that can fulfill the desired intent. The processed request is sent to the skill back-end, consisting of a part of the Skill Server block running on the local machine and exposed to the internet through an HTTPS tunnel, which has the task of converting the incoming request into information understandable by the decision algorithm (e.g. desired speed, name of the required task or other information on the job status) and send them through the network to the Decision Layer; the latter, after processing them, will build the response to be returned to the user and send it back to the Communication Layer.

\vspace{-2mm}
\section{Control and Prediction Layer}\label{sec:control}

    \subsection{Control Layer}\label{subsec:control layer}
    The control layer allows the implementation of multiple tasks onto the robot, both safety-related and application dependant, expressed in the form of time-varying constraints. The nature of this formulation is suitable for highly dynamic environments which require a strong degree of flexibility of the application, as it is often the case in HRC. The control architecture makes use of the techniques presented in \cite{Notomista2018CoRR} and \cite{Notomista2019CoRR} for encoding the desired behaviours of the robot as dynamic constraints onto the control input. These constraints are then enforced onto the robot by leveraging Control Barrier Functions (CBF) \cite{Ame19ecc}, following the resolution proposed in \cite{notomista2020settheoretic} and \cite{benzi2021optimal}.
    
    We consider a fully actuated velocity controlled $n$-DOFs manipulator represented by the following kinematic model:
    \vspace{-2mm}
    \begin{equation}\label{eq: robot_kin_model}
        \dot{x} = J(q)u
        \vspace{-2mm}
    \end{equation}
    in which  $x \in \mathbb{R}^{m}$ is the pose of the end-effector and $q \in \mathbb{R}^{n}$ is the vector of joint variables. $J(q) \in \mathbb{R}^{m \times n}$ is the Jacobian of the robot and $u \in \mathbb{R}^{n}$ is the joint velocity input.\\
    We take into account tasks whose execution can be encoded as the minimization of a non-negative, time-varying, continuously-differentiable cost function $C : \mathbb{R}^{n} \times \mathbb{R} \rightarrow \mathbb{R}$. Using the robot model in \eqref{eq: robot_kin_model} and considering as an output variable the time-varying task variable $\sigma \in \mathbb{R}^{n}$, we can represent it by means of the following optimization problem:
    \vspace{-2mm}
    \begin{equation} \label{eq: prob_cost_function}
        \begin{aligned}
            & \underset{u}{\text{minimize}}
            & & C(\sigma,t) \\
            & \text{subject to}
            & & \dot{x} = J (q)u \\
            &&& \sigma = k(x,t)
        \end{aligned}
        \vspace{-1mm}
    \end{equation}
    
    We can then utilize CBFs for encoding this problem as a convex one, for a computationally efficient resolution.
    We define the subset $\mathcal{C}\subset \mathbb{R}^n$  as the satisfaction region of the task, i.e. where $C(\sigma,t)=0$ and positive everywhere else. Consider the control barrier function $h:\mathbb{R}^{n} \times \mathbb{R} \rightarrow \mathbb{R}$ specified as $h(\sigma, t)=-C(\sigma,t)$. By design, $h$ is non-negative only in the region where the task is considered executed, i.e. when $C(\sigma,t)=0$. In this way, by enforcing the non negativity of $h$ we can achieve the execution of the task $\sigma$. This can be formulated as a convex optimization problem:
    \vspace{-1mm}
    \begin{equation} \label{eq: prob CBF general}
    \begin{aligned}
    & \underset{\dot{q}}{\text{minimize}}
    & & ||\dot{q}||^{2} \\
    & \text{subject to}
    & & \frac{\partial h}{\partial t} + \frac{\partial h}{\partial \sigma}\frac{\partial \sigma}{\partial x}J(q)\dot{q} + \alpha(h(\sigma,t)) \geq 0
    \end{aligned}
    \end{equation}
    where $\alpha(\cdot)$ is an extended class $\mathcal{K}$ function\footnote{An extended class $\mathcal{K}$ is a function $\phi:\mathbb{R}\rightarrow\mathbb{R}$ such that $\phi$ is strictly increasing and $\phi (0)=0$} and where we have considered $u=\dot q$ as the input of \eqref{eq: robot_kin_model}. Furthermore, as shown in \cite{ames2019control}, \eqref{eq: prob CBF general} makes $\mathcal{C}$ asymptotically stable, i.e. the system is attracted towards the region of satisfaction of the task.

    This formulation is intrinsically modular and lends itself easily to the inclusion of multiple tasks in the stack, i.e. the skills of the robot. Given a set of M different tasks $T_{1}, \dots , T_{M}$ which have to be executed, each respectively encoded using the cost functions $C_{1}, \dots, C_{M}$, they can be simultaneously implemented by solving the following convex optimization problem:
    \begin{equation} \label{eq: prob CBF multitask slack}
        \begin{aligned}
            & \underset{\dot{q}, \delta}{\text{minimize}}
            & & ||\dot{q}||^{2} + l||\delta||^{2} \\
            & \text{subject to}
            & & \frac{\partial h_{m}}{\partial t} + \frac{\partial h_{m}}{\partial \sigma}\frac{\partial \sigma}{\partial x}J(q)\dot{q} \\
            &&& + \alpha (h_{m}(\sigma,t)) \geq -\delta_{m} \quad  m \in  \{ 1, \dots, M\}
        \end{aligned}
    \end{equation}
    in which $h_{m}(\sigma,t) = -C_{m}(\sigma,t)$ and $\delta = [\delta_{1},\dots,\delta_{M}]^{T}$ is the vector of slack variables dedicated to each constraint, while $l \geq 0$ is a scaling factor. Each $\delta_{m}$ represents the degree of relaxation of the $m-$th task. In this way, the feasibility of the problem is ensured even in the presence of conflicting constraints. Through this formulation, the control layer is capable of implementing the actions requested by the decision layer, by encoding the received commands as tasks in the stack with appropriate parameters. Thus, using \eqref{eq: prob CBF multitask slack}, it is possible to reproduce, even simultaneously, multiple skills of the robot.
    
    Alongside the tasks commanded by the operator, a set of additional constraints of different nature can be inserted in the stack. These can be related to both safety restrictions, e.g. maintaining a certain separation distance from the human or from detectable obstacles in the scene, both to application limitations, e.g. respecting the maximum joint velocities and accelerations of the robot, its joint limits and the maximum outreach of the manipulator. For each constraint, it is possible to find a region of satisfaction $\mathcal{P}_i\subset \mathbb{R}^m$, $i=1,\dots , P$, and to define a corresponding CBF $h_c(x,t)$ that is positive on $\mathcal{P}_i$, zero on its border and negative outside of $\mathcal{P}_i$. The constraints satisfaction can be embedded in the optimization problem \eqref{eq: prob CBF multitask slack} (see \cite{ames2019control} for more details), whose final formulation is
    \begin{equation} \label{eq: prob_CBF_constrants_final}
        \begin{aligned}
             \underset{\dot{q}, \delta}{\text{minimize}}
             &\quad||\dot{q}||^{2} + l||\delta||^{2} \\
            \text{subject to}
             &\quad\frac{\partial h_{m}}{\partial t} + \frac{\partial h_{m}}{\partial \sigma}\frac{\partial \sigma}{\partial x}J(q)\dot{q}  \quad \mathbf{(S)}\\
            &\quad + \alpha (h_{m}(\sigma,t)) \geq -\delta_{m} \quad  m \in  \{ 1, \dots, M\}\\
            &\quad \frac{\partial h_{c}}{\partial t} + \frac{\partial h_{c}}{\partial x}J(q)\dot{q}  \quad \mathbf{(L)}\\
            &\quad + \alpha (h_{c}(x,t)) \geq 0 \quad  c \in  \{ 1, \dots, P\}
        \end{aligned}
    \end{equation}

    
    The optimization problem \eqref{eq: prob_CBF_constrants_final} encodes a set of tasks, i.e. the skills of the robot, that can be requested to the robot and a set of constraints, i.e. the limits of the robot, that need to be satisfied. While some flexibility in the execution of the skills is left to the robot, i.e. slack variables on the $\mathbf{(S)}$ constraints, no flexibility is left for the limitations, i.e. the $\mathbf{(L)}$ constraints have always to be satisfied. In this way, \eqref{eq: prob CBF general} will produce a control action that implements the requested skills at its best while respecting the robot limitations. Furthermore, thanks to the modularity of CBFs, it is possible to dynamically add/eliminate the skills/limits  by simply adding/eliminating the corresponding constraints. This provides great flexibility in the execution of the tasks. Finally, notice that, following the presented procedure, all the encoded constraints, as well as the objective function, result convex. Thus, the overall optimization problem results convex, allowing for a computationally fast resolution.

    
    \subsection{Prediction module}\label{subsec:prediction module}
    Before being physically implemented by the robot, each action is first transmitted to the prediction module. This module uses the information on the current condition of both the robot and the scene to simulate the execution of the desired behaviour in a virtual environment. The action is considered feasible if all the multiple tasks it imposes can be implemented simultaneously in a finite time window.

    During the simulation, the robot moves in a virtual scene by solving the problem \eqref{eq: prob_CBF_constrants_final}  and applying the obtained control input on \eqref{eq: robot_kin_model}.
    If the desired tasks $C_i(\sigma, t)=-h_m(\sigma,t)$, $m=1,\dots,M$, can be achieved,  i.e. $|h_m(\sigma, t)|<\varepsilon$, where $\varepsilon > 0$ is a desired tolerance, after a possible transient, the command is forwarded to the control layer which proceeds to the implementation of the desired behavior. 
    
    If the action is not feasible, i.e. $|h_m(\sigma,t)|\geq \varepsilon$ after the time window, or if some of the constraints encoding the limits of the robot are violated, it is necessary to look for alternatives, if any. To this aim, the limits of the robot that can be made milder are relaxed. Such limits are, e.g., minimum distance from obstacle but not strict ones such as joint limits. Suppose that, without loss of generality, the first $0<Q<P$ constraints related to the limits of the robot are the relaxable ones and let $\eta = [\eta_{1},\dots,\eta_{Q},0,\dots,0]^T$, $\eta \in \mathbb{R}^P$ the vector of slack variables related to the limits constraints. In order to generate a viable alternative the following optimization problem is solved:
    \begin{equation} \label{eq: prob_CBF_constrants_veryfinal}
        \begin{aligned}
             \underset{\dot{q}, \delta}{\text{minimize}}
             &\quad||\dot{q}||^{2} + l(||\delta||^{2} + ||\eta||^2) \\
            \text{subject to}
             &\quad\frac{\partial h_{m}}{\partial t} + \frac{\partial h_{m}}{\partial \sigma}\frac{\partial \sigma}{\partial x}J(q)\dot{q}  \quad \mathbf{(S)}\\
            &\quad + \alpha (h_{m}(\sigma,t)) \geq -\delta_{m} \quad  m \in  \{ 1, \dots, M\} \\
            &\quad \frac{\partial h_{c}}{\partial t} + \frac{\partial h_{c}}{\partial x}J(q)\dot{q}  \quad \mathbf{(L)}\\
            &\quad + \alpha (h_{c}(x,t)) \geq -\eta_c \quad  c \in  \{ 1, \dots, P\}
        \end{aligned}
    \end{equation}
    
    The simulation is run  by solving the problem \eqref{eq: prob_CBF_constrants_veryfinal} and applying the obtained control input on \eqref{eq: robot_kin_model}. If, despite of the relaxation, the requested tasks remain unfeasible, then the reason of the infeasibility is communicated to the operator via the decision layer, waiting for acknowledgment.
    Otherwise, if the relaxation is successful and the requested tasks can be accomplished, the prediction module sends to the decision layer an infeasibility message together with a possible alternative (i.e. which constraints are active and how much they need to be relaxed for succeeding). Thus, exploiting the formulation in \eqref{eq: prob_CBF_constrants_veryfinal} it is possible to make the robot aware of its skills and limits and proactive in the generation of possible alternatives to propose to the human operator. The operator can either accept or reject the proposed alternative.

\section{Decision Layer}\label{sec:decision}
\begin{figure} [htbp]
        \centering
        \includegraphics[trim=0cm 0cm 0cm 0cm,clip,scale=0.5]{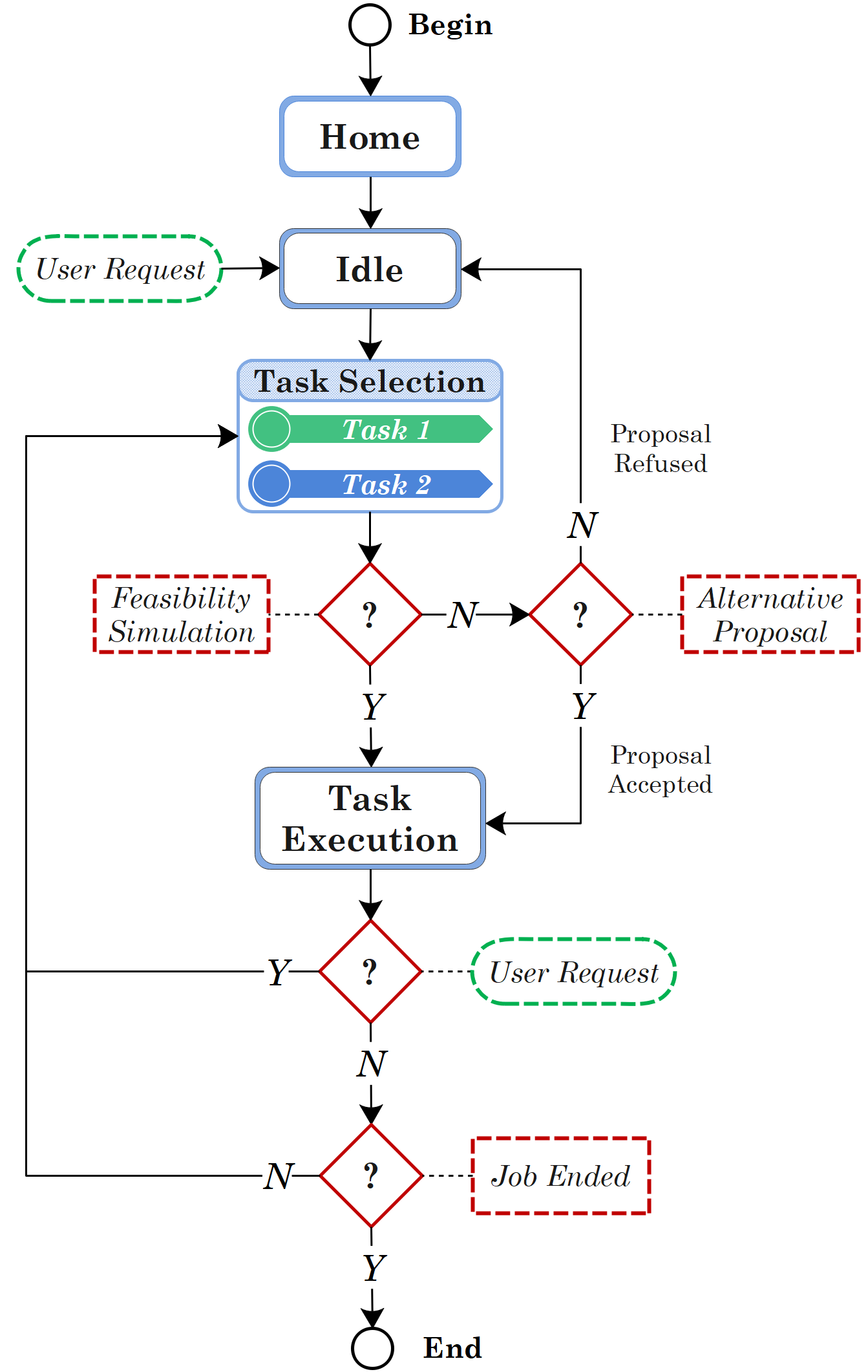}
        \caption{Decision Layer Flowchart}
        \label{fig:decision_layer}
    \end{figure}

    The decision layer has the task of collecting all the information coming from the communication and the control layer and comparing them to make a decision on the next action to be carried out. The layer is structured using a sequential flow chart architecture, as shown in Fig. \ref{fig:decision_layer}, defining the decision-making process that occurs during the execution of the HRC job. 
    When the operator makes a request to the robot, the decision layer evaluates the current state of the robot, the extent of the request and communicates to the control layer the willingness to carry out this action. The latter simulates the action and provides an answer on the feasibility of the requested action. The decision layer will then build a sentence to communicate the result of the processing to the user and, in case of feasibility, it will start the execution of the task, otherwise it will communicate to the user the failure of the simulation and can propose alternative solutions. The alternatives are elaborated on the basis of the feedback coming from the control layer and subsequently proposed to the user through the communication layer. If the user accepts the alternative proposal, the decision layer replaces the previously selected task with the alternative proposed one and proceeds with the execution; if the proposal is rejected, the decision layer goes back to the “IDLE” status waiting for a new command to be able to continue the job.
    

    When the job starts, the robot first moves to the “HOME” position and then enters the “IDLE” state waiting for a command. Depending on the received user request, one of the available tasks is selected and the decision layer starts executing it. First, by communicating with the control layer, the movement feasibility check is carried out and, if necessary, alternatives are worked out and proposed to the user. Once the execution is completed, the decision layer checks if there are any operator requests in the queue and, if there are none, it checks if the job has come to the end: if it has not yet finished, the job resumes selecting the next task to be performed, otherwise the robot stops and communicates the end of the job, waiting for a new start.
    
\section{Experimental Validation}\label{sec:exp}

    The experimental validation was carried out with a comparative experiment in which bidirectional communication control was compared with the common one-way communication control. The two experiments were performed in a set of 12 participants, aged between 20 and 46 years, randomly extracting the order of execution of the experiments to limit possible effects of novelty and learning during the experiments. The choice of participants was restricted due to COVID-induced limitations to the personal belonging to the laboratory, i.e. persons with an engineering background but different levels of expertise in robotics.
    The considered scenario is a highlight of a human-robot collaborative assembly job in a shared workspace, during which an obstacle is placed by the operator in the task operation area of the robot, a UR10e 6DoF collaborative manipulator \cite{UR10e}. The robot has to complete a series of pick and place tasks but the presence of the obstacle, represented by a TurtleBot3 Burger robot \cite{turtlebot} that is the result of the collaborative assembly, prevents it from finishing the operations.
    The use of the bidirectional communication control architecture allows the robot to verify the feasibility of the task, via the embedded simulator, and to eventually elaborate and propose to the operator possible alternative solutions. This is carried out by dynamically modifying specific control parameters, considering the limits and skills of the robot and the environmental constraints.

    \subsection{Architecture Implementation}\label{subsec:implementation}
        
\begin{figure} [htbp]
            \centering
            \includegraphics[scale=0.05]{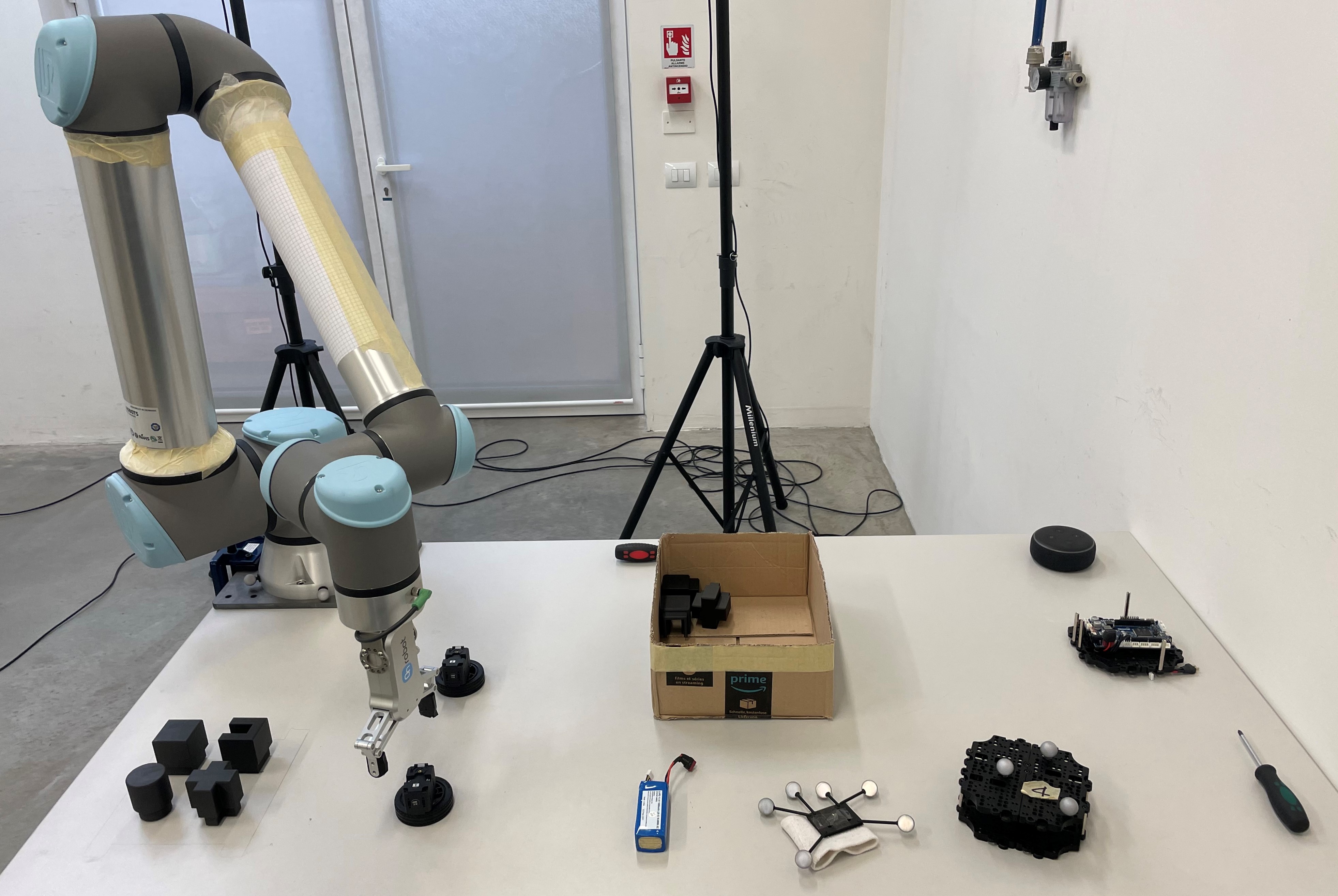}
            \caption{Setup of the Experiment}
            \label{fig:experiment_setup}
            \vspace{-2mm}
        \end{figure}

        The architecture was built using ROS \cite{ROS}, dividing the various layers into independent nodes in order to ensure the modularity of the control architecture and make it compatible with multiple communication systems and decision making algorithms. 
        The communication layer consists of a voice communication channel, implemented by connecting via HTTPS tunnel a custom skill, based on the Alexa Voice Service (AVS) ecosystem \cite{AVS} and performed on an Alexa Echo Dot commercial voice assistant, with a server running on the local computer that acts as the endpoint of the skill. This server is written using Flask-Ask \cite{Flask-Ask}, an extension of the Flask framework \cite{Flask} that makes building Alexa skills easier and faster, and Flask-ngrok \cite{Flask-ngrok} that allows to integrate the HTTPS tunnel, created through the ngrok software \cite{ngrok}, directly into the local server. In addition, a Text-To-Speech node has been added by integrating into ROS and optimizing the “ha-alexa-tts” shell scrip \cite{ha-alexa-tts}.
        Decision and Control layers also run on the local computer and communicate with each other through the ROS ecosystem. To allow the control layer to derive information from the environment we choose the OptiTrack real time 6DoF tracking systems \cite{optitrack}. The setup is shown in Fig. \ref{fig:experiment_setup}.
        
    \subsection{Skills and Constraints}\label{subsec:cbf_task}
        The robot is set to accomplish multiple position tasks while satisfying a set of constraints, both safety-related, both application-related, whose implementation is achieved by synthesizing the control input through \eqref{eq: prob CBF multitask slack}.
        
        The stack of constraints includes obstacle avoidance and respect of the joint limits. Both tasks and constraints are encoded by means of an appropriately designed CBF, formulated via the procedure exposed in Sec.~\ref{sec:control}.\\
        As an example, the obstacle avoidance constraint is encoded through the following CBF:
        \begin{equation}\label{eq: safety CBF}
            h_{safe} = -\rho_{s}(d^{2} - D_{min}^{2})
        \end{equation}
        in which $\rho_{s} = 10$ is a tuning gain, $D_{min}$ is the minimum distance value between the obstacle and the tip of the end-effector, while $d = d(x, t)$ is defined as
        \begin{equation}\label{eq: distance definition}
            d = ||x - x_{obs}||^{2}
        \end{equation}
        where $x_{obs}$ is the Cartesian position of the nearest obstacle.\\
        According to the procedure in \eqref{eq: prob CBF multitask slack}, for each CBF the gain $\rho_{(.)}$ is tuned and, after simple computations, the corresponding constraint is derived and inserted into the optimization problem. For the sake of simplicity, the function $\alpha$ was chosen as the identity function, such that $\alpha(h_{m}(\sigma,t)) = h_{m}(\sigma,t) \, \forall m \in \{ 1,\dots, M\}$. From \eqref{eq: safety CBF} following the formulation in \eqref{eq: prob CBF multitask slack}, we derive the following constraint:
        \begin{equation}\label{eq: safety CBF constraint}
            2(d - D_{min})J(q)\dot{q} \geq -(h_{safe}(x,t)) + \delta_{s}
        \end{equation}
        in which $\delta_{s}$ is the dedicated slack variable.
        
        The joint limits constraints maintain the position of each joint within an upper and lower limit. Each $i-$th joint presents a dedicated CBF, implemented as:
        \begin{equation}\label{eq: joint limit CBF}
             h_{lim_{i}} = \rho_{l}\frac{(q^{+}_{i} - q_{i})(q_{i}- q^{-}_{i})}{(q^{+}_{i} - q^{-}_{i})} \quad i \in  \{ 1, \dots, n\}
        \end{equation}
        in which $\rho_{l} = 1$, while $q^{+}_{i}$ and $q^{-}_{i}$ indicate the real joint limits of the $i$-th joint, in the joint space. Notice how keeping the value of $q_{i} \in [q^{+}_{i}, q^{-}_{i}]$ can be achieved by imposing $h_{lim_{i}} \geq 0$.
        Finally, the position control task is encoded through the following CBF:
        \begin{equation}\label{eq: pos control CBF}
            h_{pos} = -\rho_{p}||x - x_{goal}||^{2}
        \end{equation}
        in which $\rho_{p} = 5$ and $x_{goal}$ is the current goal, expressed as the desired Cartesian position of the end effector.

    \subsection{Analysis of the Results}\label{subsec:results}

        To verify the effectiveness of the architecture we evaluated two different fundamental aspects:

        \begin{itemize}
            \item The system's ability to complete all required tasks.
            \item The differences in the execution time of the operations.
        \end{itemize}

\begin{figure} [htbp]
            \centering
            \includegraphics[trim=2.1cm 10cm 2.6cm 12.7cm,clip,scale=0.5]{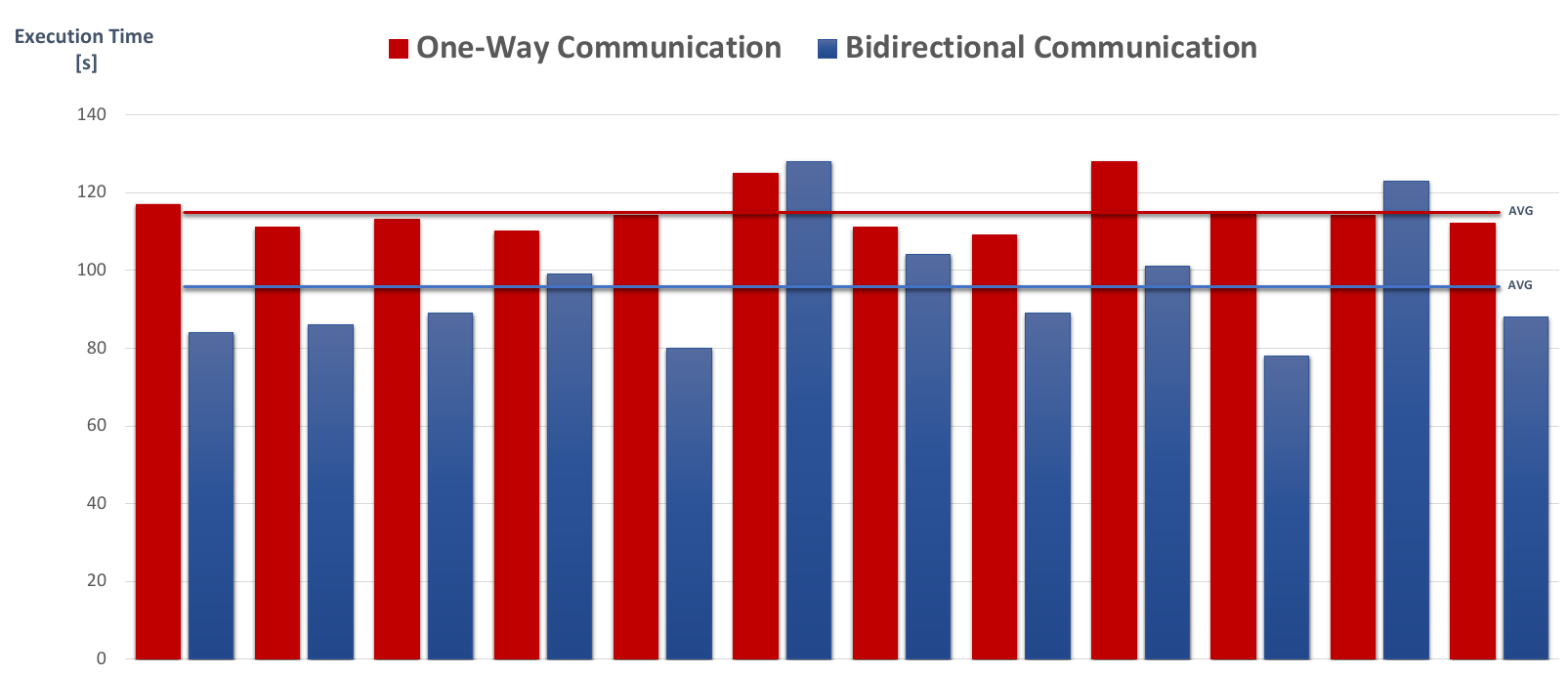}
            \caption{Execution Time Comparative Results. The horizontal lines indicates the average and the columns are the time measurements of the job.}
            \label{fig:time}
            \vspace{-0.2cm}
        \end{figure}

        From the analysis of the collected data, it was evident that the use of the bidirectional communication architecture can bring noticeable advantages to the execution of the requested operations. While using normal one-way communication the presence of an obstacle often led to the abortion of the task, while with the proposed architecture, owing to its ability to simulate and propose possible alternatives, the system was able to adapt and perform the desired task anyway (i.e. relax a minimum distance constraint from the obstacle and moderating the execution speed of the task).
        Considering therefore that in some experiments it was not possible to complete all the tasks, it was necessary to introduce a penalty function $P = \kappa t_m$ (in which $\kappa$ is the penalty coefficient and $t_m$ the average execution time of the failed task) to the execution time of the experiment in case the system remained stationary without providing any feedback to the operator and therefore considering that the requested job not fully completed.
        In order to assign a value to the penalty coefficient that reflects the inconvenience that the failure of the requested task entails, two main aspects were evaluated:
        \begin{itemize}
            \item The time lost for the operator to understand the reason for the error, a time-factor that grows with the length and complexity of the involved operations, and elaborates a possible solution to the problem.
            \item The total time to request the robot again to perform the failed operation and wait for the execution, is certainly higher than the average execution time of the task.
        \end{itemize}
        Following these considerations, it was evaluated that a $\kappa$ value equal to 1.5 reflects the inconvenience caused by the failure of the task. 
        The graph in figure \ref{fig:time} shows the results of the measurements of the experiment execution time: in red the data concerning the one-way communication, in blue the bidirectional communication. We can observe that the use of the proposed architecture involves a significant improvement in performance of the job, taking on average 96 seconds to complete the operations, against the average 115 for one-way communication.
        
        
         To better assess the statistical validity of the experiment and confirm the conclusions drawn, we performed a single factor ANOVA test. Table \ref{tab:ANOVA_summary_results} shows the summary values of the collected data (mean, sum and variance) divided into the two experiments and the obtained results: we can observe that the calculated F-value is significantly greater than the F-critical and the p-value is much less than the significance level (alpha = 0.05), thus confirming a statistical difference between the two experiments.
         
        
        
        \begin{table}[htbp]
            \caption{ANOVA Summary and Results}
            \label{tab:ANOVA_summary_results}
            \centering
            \resizebox{\linewidth}{!}{
            \begin{tabular}{c|c|c|c|c}
                \toprule
                \textbf{Groups} & \textbf{Count} & \textbf{Sum} & \textbf{Average} & \textbf{Variance} \\
                \midrule
                One-Way Communication & 12 & 1379 & 114.917 & 34.629 \\
                Bidirectional Communication & 12 & 1149 & 95.75 & 257.841 \\
                \bottomrule
                \toprule
                \textbf{F} & \multicolumn{2}{c|}{\textbf{P-value}} & \multicolumn{2}{c}{\textbf{F crit.}} \\
                \midrule
                15.073 & \multicolumn{2}{c|}{0.00080} & \multicolumn{2}{c}{4.30095} \\
				\bottomrule
            \end{tabular}}
        \end{table}
        
        
    
\vspace{-1mm}
\section{Conclusions and Future Work}\label{sec:conclusions}

    In this paper, we have proposed a bidirectional human-robot communication architecture that allows a human-robot team to collaborate during the execution of a shared task by exchanging information in a bidirectional way in order to achieve the intended objectives. Through this constant exchange of information regarding the external environment, the status of the operator or robot task, requests or proposals from one of the two team members, we have achieved a significant improvement in the effective ability to complete the tasks required even with unforeseen situations.
    As further extensions of this architecture, we evaluated the idea of adding further communication channels to expand the ability of the system to interact with the operator. Furthermore, a future goal is to move the control in the Cartesian space, in order to obtain more precise and more legible robot movements. We also plan to improve the artificial intelligence of the decision layer, introducing algorithms of teach by demonstration and machine learning to give the operator the possibility to teach the robot how to perform the required tasks and how to react to particular situations.
    

\bibliographystyle{IEEEtran}
\bibliography{mybib.bib}

\end{document}